\documentclass[runningheads]{llncs}

\usepackage[T1]{fontenc}

\usepackage{graphicx}
\makeatletter
\renewcommand\subparagraph{\paragraph}
\makeatother
\usepackage[labelfont=bf,labelsep=colon]{caption} %
\pdfpageattr{/Group << /S /Transparency /I true /CS /DeviceRGB>>}
\usepackage[hypcap=false]{caption}
\usepackage{subfig} 
\usepackage{float}  
\usepackage{placeins} 

\usepackage{multirow}
\usepackage{booktabs}
\usepackage{overpic}   
\usepackage{booktabs}
\usepackage{amsmath}
\usepackage{amssymb}
\usepackage{subcaption} 
\usepackage[colorlinks=true, linkcolor=blue, urlcolor=blue, citecolor=blue]{hyperref}
\usepackage{graphicx}
\usepackage[utf8]{inputenc}  
\usepackage[T1]{fontenc}     
\usepackage{verbatim}

\begin{document}
\title{Text-Guided Multi-Instance Learning for Scoliosis Screening via Gait Video Analysis}

\author{
  Haiqing Li \and
  Yuzhi Guo \and
  Feng Jiang \and
  Thao M. Dang \and
  Hehuan Ma \and
  Qifeng Zhou \and
  Jean Gao \and
  Junzhou Huang*
}

\authorrunning{H. Li et al.}
%
\institute{Department of Computer Science and Engineering, The University of Texas at
Arlington, Arlington, TX 76019, USA \\ 
\email{jzhuang@uta.edu}}
\maketitle              
\begin{abstract}
Early-stage scoliosis is often difficult to detect, particularly in adolescents, where delayed diagnosis can lead to serious health issues. Traditional X-ray-based methods carry radiation risks and rely heavily on clinical expertise, limiting their use in large-scale screenings. To overcome these challenges, we propose a \textbf{T}ext-\textbf{G}uided \textbf{M}ulti-\textbf{I}nstance \textbf{L}earning Network (TG-MILNet) for non-invasive scoliosis detection using gait videos. To handle temporal misalignment in gait sequences, we employ Dynamic Time Warping (DTW) clustering to segment videos into key gait phases. To focus on the most relevant diagnostic features, we introduce an Inter-Bag Temporal Attention (IBTA) mechanism that highlights critical gait phases. Recognizing the difficulty in identifying borderline cases, we design a Boundary-Aware Model (BAM) to improve sensitivity to subtle spinal deviations. Additionally, we incorporate textual guidance from domain experts and large language models (LLM) to enhance feature representation and improve model interpretability. Experiments on the large-scale Scoliosis1K gait dataset show that TG-MILNet achieves state-of-the-art performance, particularly excelling in handling class imbalance and accurately detecting challenging borderline cases. The code is available at \url{https://github.com/lhqqq/TG-MILNet}

\end{abstract}
\section{Introduction}
\label{sec:intro}
Scoliosis, an abnormal spinal curvature, is often difficult to detect in its early stages. Most cases in children and adolescents are idiopathic, with no identifiable cause~\cite{ref1,ref2}. Without timely intervention, scoliosis can progress and significantly affect health and quality of life. Adolescent scoliosis affects a significant portion of youth worldwide, with increasing incidence in China in recent years~\cite{hlj2024}.

\begin{figure}[htbp]
        \centering
        \includegraphics[width=0.8\linewidth]{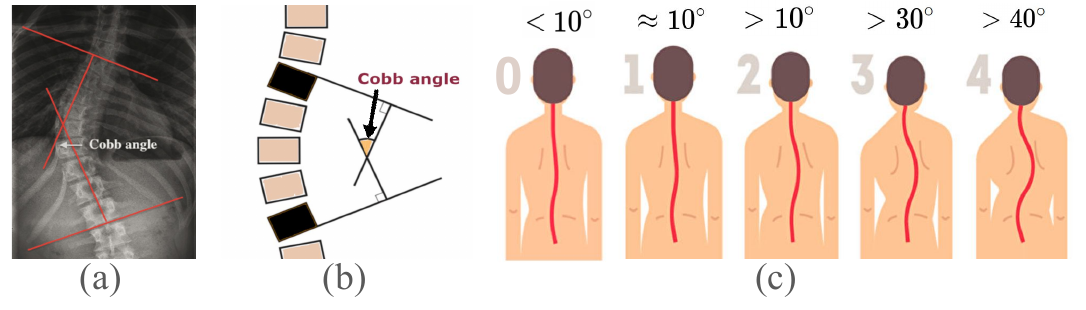}
        \caption{Illustrations of Cobb angle calculation for scoliosis assessment: (a) X-ray measurement, (b) schematic representation, and (c) scoliosis progression..}
        \label{fig:cobb}
    \end{figure}

Scoliosis diagnosis relies on measuring the Cobb angle, as shown in Fig.\ref{fig:cobb}. Scoliosis is diagnosed when the Cobb angle exceeds 10 degrees~\cite{ref4}. However, X-rays require clinical expertise and pose radiation risks, making them impractical for large-scale screening. Deep learning–based approaches, like back photo analysis~\cite{ref6}, provide non-contact alternatives but struggle with privacy concerns and patient compliance~\cite{ref7,radford2021learning,ref9}. To overcome these limitations, ScoNet-MT\cite{ref10} introduces gait patterns as biomarkers for scoliosis detection and provides Scoliosis1K, the first video-based gait dataset for scoliosis. It encodes silhouette sequences frame-by-frame, aggregates representations, and predicts scoliosis. Given the dynamic nature of gait, not all frames contribute equally to the representation of gait patterns. Multiple Instance Learning (MIL)\cite{ref11} and Graph learning\cite{jiang2024alphaepi,jiang2024gte} mitigate this issue by selectively emphasizing diagnostic frames. 

Despite these advancements, distinguishing borderline cases remains difficult, as subtle spinal deviations near the 10-degree threshold are often misclassified as normal. Borderline cases indicate early scoliosis, so timely, accurate detection is critical, as shown in Fig.~\ref{fig:cobb}(c). In addition, class imbalance increases the risk of missed borderline cases, highlighting the need to enhance sensitivity to mild deformities.~\cite{weinstein2008adolescent,cheng2015adolescent}.

Inspired by prior work~\cite{ref10,li2025leveraging}, we propose TG-MILNet, a multimodal and multi-instance learning framework for scoliosis detection. Like other biometrics~\cite{wang2023metascleraseg,wang2024sclera,ref5}, gait exhibits inherent consistency and discriminative characteristics, making it a promising biomarker for health screening. It is designed to effectively capture and analyze discriminative gait patterns from video data. To handle the temporal misalignment and variability in gait sequences, we introduce a Dynamic Time Warping (DTW)-based clustering method~\cite{petitjean2011global} to splits gait videos into distinct phases (e.g., front view, turning, back view) capturing both temporal and spatial dynamics. Each cluster serves as a bag of instances for MIL, providing structured input for downstream feature extraction. Gait phases contribute inconsistently to scoliosis detection, making it crucial to model their relationships effectively. To address this, we propose the Inter-Bag Temporal Attention (IBTA) mechanism, which dynamically emphasizes informative gait phases while filtering out irrelevant variations. This enables the model to focus on patterns most indicative of scoliosis. Borderline cases and class imbalance present significant challenges in classification. To address these issues, we introduce the Boundary-Aware Model (BAM), an auxiliary module designed to distinguish borderline cases from other samples, thereby improving classification accuracy in difficult scenarios. To enhance interpretability and guide feature learning, we incorporate textual guidance derived from expert domain knowledge~\cite{cheng2015adolescent} and GPT-4o~\cite{achiam2023gpt}-generated insights. These textual cues help the model concentrate on scoliosis-related gait abnormalities, improving robustness and interpretability. We validate TG-MILNet on Scoliosis1K, the first large-scale video-based scoliosis gait dataset, demonstrating its effectiveness in early-stage scoliosis detection and its potential as a scalable screening tool.

In summary, our key contributions include: 1) the proposal of \textbf{TG-MILNet}, a multimodal and multi-instance learning framework for scoliosis detection using gait videos; 2) a \textbf{DTW-based clustering} method to cluster gait sequences into key phases and address temporal misalignment; 3) the \textbf{Inter-Bag Temporal Attention (IBTA)} mechanism to model relationships between gait phases and emphasize informative frames; 4) the \textbf{Boundary-Aware Model (BAM)} to improve classification accuracy for borderline cases and mitigate data imbalance; and 5) the integration of \textbf{textual guidance} from expert knowledge and GPT-4o-generated insights to enhance model interpretability and robustness.

\section{Methodology}
\label{sec:Methodology}

\subsection{Dynamic Time Warping-based clustering}
\label{Clustering}

\begin{figure}[t]
    \centering
    \includegraphics[width=0.88\textwidth]{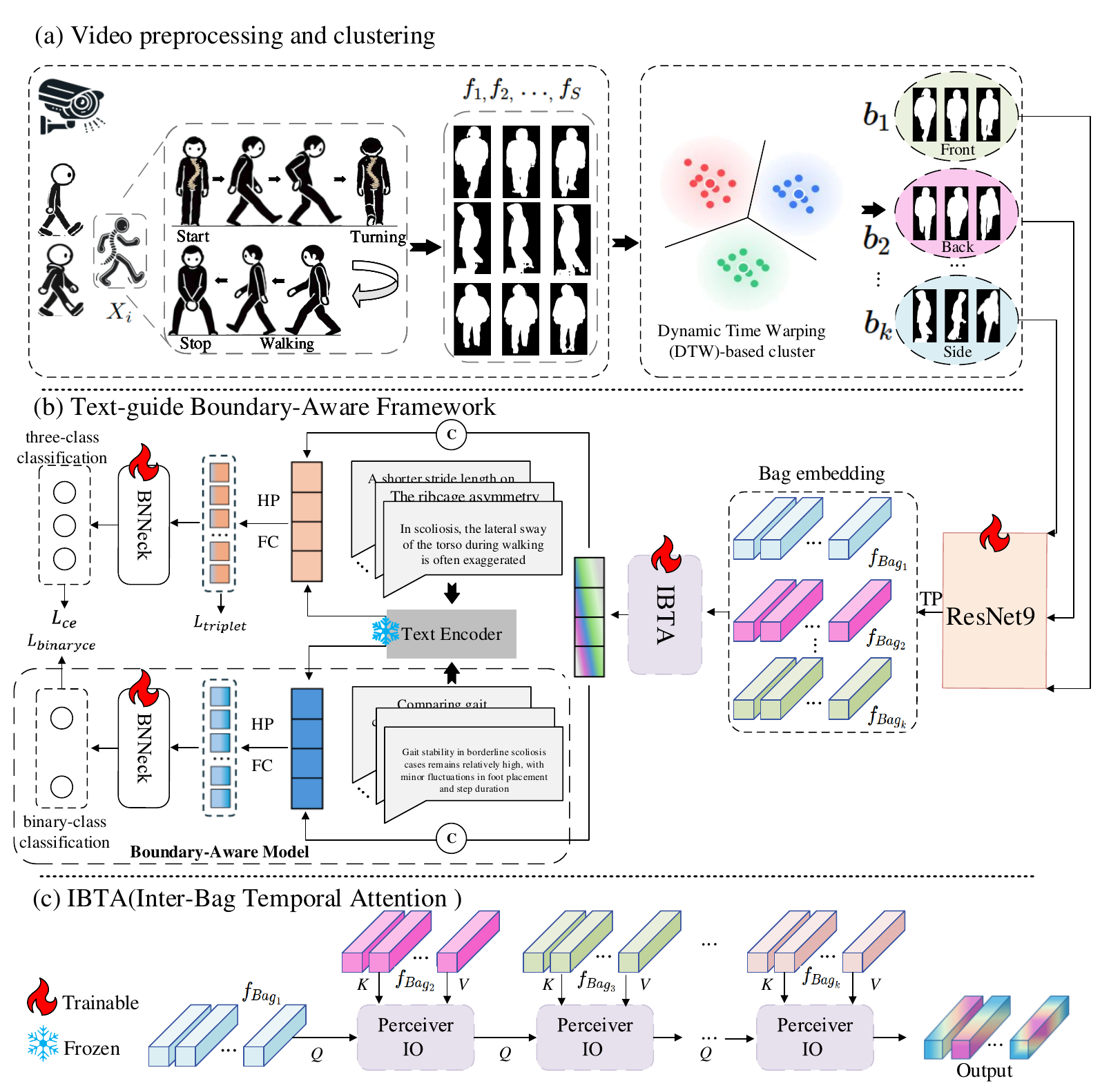} 
    \caption{\small The TG-MILNet flowchart includes: (1) DTW-based clustering that segments video frames into temporal bags, enabling phase-specific feature extraction, (2) IBTA mechanism that prioritizes informative gait phases and filters irrelevant variations, and (3) a dual-branch classification framework addressing data imbalance and borderline cases. Expert and GPT-4-based textual guidance enhances interpretability by emphasizing scoliosis-related gait patterns.} 
    \label{fig:overview}
\end{figure}

The overview of TG-MILNet is shown in Fig.~\ref{fig:overview}. 
As illustrated in Fig.~\ref{fig:overview}(a), a gait video is preprocessed by extracting frames and clustering them into phases. 
Each subject \( X_i \) has a sequence of gait frames \(\{ f_{1}, f_{2}, \dots, f_{S} \}\), where \( S \) is the total number of frames. 
To handle temporal misalignment, we use a Dynamic Time Warping (DTW)-based clustering method.  Specifically, we first extract frame-wise optical flow features and compute the DTW distance between any two frames:
\(
D(i, j) = \text{DTW}(f_i, f_j).
\)
Hierarchical clustering is then applied to these distances to group frames into \( K \) clusters (bags), denoted as \(\{ b_{1}, b_{2}, \dots, b_{K} \}\). 
Each cluster corresponds to a specific gait phase (e.g., start, mid-walk, end) or perspective (e.g., front view, turning, back view), as shown in Fig.~\ref{fig:overview}(a). This structure facilitates the subsequent analysis of movement patterns. By identifying distinct phases and viewpoints, our approach helps extract robust features that are invariant to temporal fluctuations and viewpoint changes.

\subsection{Inter-Bag Temporal Attention mechanism }
\label{IBTA}

Each bag, defined as \( \text{b}_i \in \mathbb{R}^{N \times C \times S_i \times H \times W} \), contains a varying number of frames and is processed through a ResNet-based feature extractor, yielding:

\begin{equation}
f_{\text{Bag}_i} = TP(F(\text{b}_i)) \in \mathbb{R}^{N \times C \times H \times W},
\end{equation}
where \( i = 1, 2, \dots, K \) and \( N, C, S, H, W \) represent batch size, channels, frames, height, and width, respectively, and \( TP \) denotes Temporal Pooling~\cite{fan2020gaitpart}.

The IBTA mechanism (Fig.~\ref{fig:overview}(c)) dynamically integrates information across temporal bags using cascaded \textit{Perceiver IO} modules~\cite{dang2024mfmf,dang2025abnormality}, ensuring efficient modeling of temporal relationships and multi-level feature fusion. At its core, the cross-attention mechanism exchanges information between input bags and progressively integrates bag embeddings \( \{f_{\text{Bag}_1}, f_{\text{Bag}_2}, \dots, f_{\text{Bag}_k}\} \) through iterative attention operations. In each block, the query (\(Q\)) is derived from the current bag, while the key (\(K\)) and value (\(V\)) come from the preceding output. For instance, the first block uses \( f_{\text{Bag}_1} \) as \(Q\), and both \(K\) and \(V\) are set to \( f_{\text{Bag}_2} \), enriching \( f_{\text{Bag}_1} \) with temporal features from \( f_{\text{Bag}_2} \). In subsequent blocks, the output \( H_1 \) becomes the new query (\(Q\)), while the key and value are derived from the next bag, continuing the refinement process:
\begin{equation}
H_1 = \text{Attention}(f_{\text{Bag}_1}, f_{\text{Bag}_2}, f_{\text{Bag}_2}),
    H_2 = \text{Attention}(H_1, f_{\text{Bag}_3}, f_{\text{Bag}_3}).
\end{equation}

This iterative mechanism allows each step to refine the learned representations by incorporating information from the next bag, progressively capturing global temporal dependencies and multi-level contextual features. The final output encodes comprehensive information from all bags.

\subsection{Text-Guided Boundary-Aware Framework }
\label{dual}

In clinical practice, a Cobb angle of 10 degrees is the diagnostic threshold for scoliosis. Borderline cases near this threshold, with subtle spinal curvatures, are often misclassified as normal due to minimal deviations and the predominance of normal patterns in real-world datasets. This highlights the dual challenge of data imbalance and the difficulty in detecting subtle abnormalities. To address this, we introduce BAM: instead of directly classifying normal, borderline, and pathological cases, BAM focus on distinguishing borderline cases from both normal and pathological samples via binary classification. To enhance gait-based visual features, we incorporate textual guidance for domain-specific knowledge. For the three-class classification, the text highlights scoliosis-related abnormalities, helping the model focus on key gait deviations (e.g., “In scoliosis, the lateral sway of the torso during walking is often exaggerated”). For the binary-class classification, it emphasizes subtle differences between borderline and other samples, improving classification of challenging cases. By incorporating text, the model gains access to expert knowledge that it might not infer solely from visual data, enabling more robust and interpretable decision-making. The textual guidance is derived from two key sources: expert domain knowledge and GPT-4o. For text features, we use the pretrained CLIP model~\cite{radford2021learning} to extract text features from the three-class and binary classification tasks. These text embeddings are then concatenated with the image-extracted features to generate a final representation that combines visual and semantic information for subsequent steps.

Horizontal Pooling (HP)~\cite{fu2019horizontal} divides the feature maps into 16 horizontal segments and applies global pooling to obtain detailed features. We further use a Separate Fully Connected layer (FC) to map features into the metric space, where the widely-used BNNeck~\cite{luo2019strong} is applied to adjust the feature space. The mapped features are optimized using the cross-entropy loss for classification tasks. We use the cross-entropy loss to optimize the classification of \textit{normal}, \textit{borderline}, and \textit{pathological} samples and specifically distinguish \textit{borderline} samples from others:
\begin{equation}
L_{\text{ce}} = - \sum_{i=1}^{n} y_i \log(\hat{y}_i), \quad L_{\text{binaryce}} = - \sum_{i=1}^{n} \left[ y_i \log(\hat{y}_i) + (1 - y_i) \log(1 - \hat{y}_i) \right],
\end{equation}
where \( y_i \) is the ground truth label and \( \hat{y}_i \) is the predicted probability for the \(i\)-th sample. To further enhance feature discrimination, we employ the triplet loss:
\begin{equation}
L_{\text{Triplet}} = \frac{1}{N_{\text{valid}}} \sum_{\substack{a,p,n \\ y_a = y_p \neq y_n}} \max\left(m + d(a, p) - d(a, n), 0\right),
\end{equation}
where \( a \), \( p \), and \( n \) represent anchor, positive, and negative samples, respectively. \( d(\cdot) \) denotes pairwise distances, \( m \) is the margin, and \( N_{\text{valid}} \) is the number of valid triplets.
The overall loss function combines these components and is expressed as:
\begin{equation}
L_{\text{Total}} = L_{\text{Triplet}} + L_{\text{ce}} + L_{\text{binaryce}}.
\end{equation}

\section{Experiments and Results}
\label{sec: Experiments}

\subsection{Dataset}
\textbf{Scoliosis1K}~\cite{ref10} is the first large-scale video-based dataset for scoliosis classification. It includes 1,050 adolescent volunteers from middle schools in China, with a total of 447,900 sequence frames. To protect privacy, all images are presented in silhouette format. The subjects are divided into three categories: \textit{positive} (Cobb angle \( > 10^\circ \)), \textit{neutral} (borderline, Cobb angle \( \approx 10^\circ \)), and \textit{negative} (Cobb angle \( < 10^\circ \)). 

\subsection{Implementation Details}

We follow the dataset’s protocol, maintaining a 1:1:8 ratio of \textit{positive, neutral, and negative}  samples in the training set~\cite{ref10}. The input resolution of dataset is set to 128 × 88 pixels. The anchor and the positive come from the same subject but consist of different frames. We experimentally demonstrate that setting the number of bags K to 4 yields the best performance. The other settings are consistent with the dataset organizers.

\subsection{Comparison with State-of-the-Art}

\begin{figure}[t]
    \centering
    \includegraphics[width=1\linewidth]{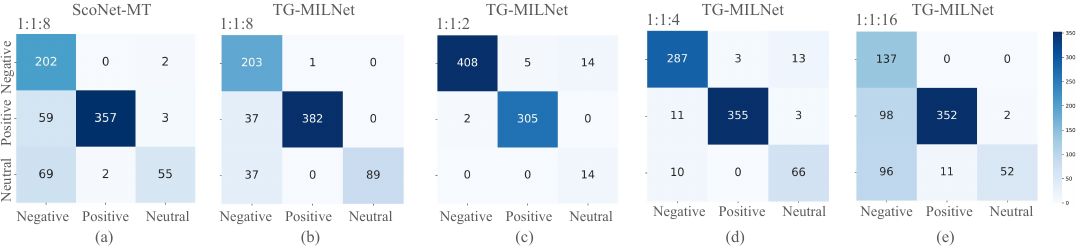}
    \caption{Confusion matrices of ScoNet-MT and TG-MILNet under different class imbalance.}
    \label{fig:cm}
\end{figure}

\begin{figure}[t]
    \centering
    \includegraphics[width=1\linewidth]{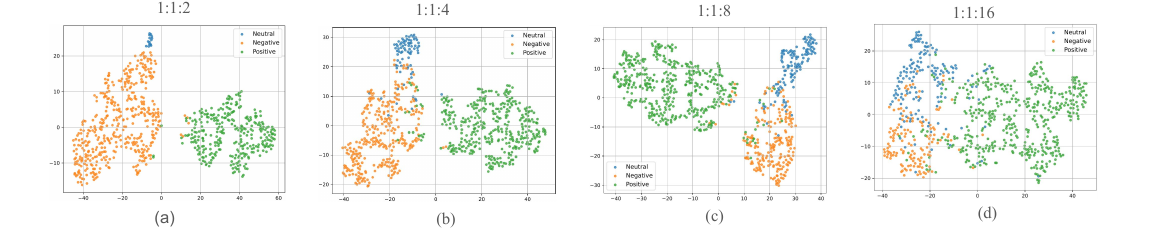}
    \caption{t-SNE Visualization on an Imbalanced Dataset.}
    \label{fig:tsne}
\end{figure}

We compare our TG-MILNet with the current state-of-the-art (SOTA) deep learning-based method ScoNet-MT~\cite{ref10}, as well as some traditional scoliosis detection techniques (as shown in Table~\ref{table: 2}). TG-MILNet surpasses existing SOTA methods, achieving the highest accuracy of 89.9\% while excelling in balancing sensitivity and specificity. These results demonstrate its superior ability to precisely  classify \textit{positive}, \textit{neutral}, and \textit{negative} samples, even under data imbalance, setting a new benchmark for scoliosis detection. Furthermore, with a sensitivity of 99.5\%, TG-MILNet demonstrates outstanding effectiveness in identifying negative samples, minimizing the likelihood of false-positive diagnoses. Notably, as shown in Table~\ref{table:dual}, TG-MILNet achieves consistent improvements across all three classes: +0.5\% (Negative), +6\% (Positive), and +27.0\% (Neutral). The neutral (borderline) class benefits the most, highlighting TG-MILNet’s superior robustness in challenging borderline scenarios where previous methods often fail. As shown in Fig.~\ref{fig:cm}~(a) and~(b), ScoNet-MT frequently misclassifies neutral cases as Negative, increasing the risk of missed diagnoses, whereas TG-MILNet effectively distinguishes these ambiguous cases, reducing misclassification and supporting more reliable screening. In addition, the t-SNE Fig.~\ref{fig:tsne} shows that the neutral class (blue dots) is well separated from the positive and negative classes, further confirming the model’s effectiveness in distinguishing subtle gait patterns. As class imbalance increases, the separation between the neutral and other classes becomes more distinct, illustrating that TG-MILNet effectively captures discriminative features and enhances class separability. TG-MILNet significantly improves the classification accuracy of small-sample positive cases under data imbalance(1:1:16). This demonstrates strong adaptability and robustness, paving the way for reliable large-scale screening in real-world settings.

\begin{table}[t]
\caption{Comparison of different methods (left) \& Ablative results on the impact of class imbalance (right). Acc, Sen, Spe $=$ accuracy, sensitivity, specificity.}
\resizebox{\textwidth}{!}{
\begin{tabular}{lcccc|l|ccccc}
\hline
\multirow{2}{*}{\textbf{Method*}} & \multicolumn{4}{c|}{\textbf{Performance (\%)}} &  & \multirow{2}{*}{\textbf{Pos:Neu:Neg}} & \multicolumn{3}{c}{\textbf{Acc (\%)}} &  \textbf{F1(\%)}  \\ \cline{2-5} \cline{8-10} \cline{11-11}
                                  & \textbf{Acc} & \textbf{Sen} & \textbf{Spe} &  \textbf{F1 }                                &  &                              & \multicolumn{1}{c}{ScoNet} & \multicolumn{1}{c}{ScoNet-MT} & TG-MILNet &  TG-MILNet \\ \hline
Adams Test       & -       & 84.4    & \textbf{95.2}    & -      &  & \multicolumn{1}{l}{1:1:2}   & 91.4  & 95.2  & \textbf{97.2}      & 97.5      \\
Scoliometer      & -       & 90.6    & 79.8    & -      &  & \multicolumn{1}{l}{1:1:4}   & 88.6  & 90.5  & \textbf{94.6}      & 94.7      \\
CNN              & 45.5    & 99.0    & 27.0    & -      &  & \multicolumn{1}{l}{1:1:8}   & 51.3  & 82.0  & \textbf{89.9}      & 90.2      \\
ScoNet           & 51.3    & \textbf{100.0}   & 33.2    & -      &  & \multicolumn{1}{l}{1:1:16}  & 23.7  & 53.2  & \textbf{72.3}      & 73.3      \\ \cline{7-11} 
ScoNet-MT        & 82.0    & 99.0    & 76.5    & 81.9   &  & \multicolumn{5}{l}{*\textit{Adams Test and Scoliometer are traditional methods,}}  \\ 
TG-MILNet (ours) & \textbf{89.9}    & \underline{99.5}     & \underline{86.4}    & \textbf{90.2}   &  &\multicolumn{5}{l}{\textit{while the others are deep learning-based methods.}}     \\ \hline
\end{tabular}
}
\label{table: 2}
\end{table}

\subsection{Ablation Studies}
\textbf{Influence of class imbalance}. To explore the effects of class imbalance in real-world, we evaluate TG-MILNet under varying \textit{positive, neutral, and negative} sample ratios (e.g., 1:1:2, 1:1:4, 1:1:16). The results are shown in Table~\ref{table: 2}, which  demonstrate that as class imbalance increases, TG-MILNet consistently outperforms baseline methods, maintaining high accuracy and stability. While its advantage is minor under mild imbalance (1:1:2) with only a 2.0\% accuracy improvement, its superiority becomes evident at 1:1:8 (+7.9\%) and 1:1:16 (+19.1\%), showcasing strong adaptability to extreme imbalance scenarios. Notably, TG-MILNet achieves higher F1-scores. This indicates the model enhances accuracy, reduces false positives, and improves minority class recognition, which is crucial for identifying borderline cases. These results highlight TG-MILNet’s robustness and effectiveness in handling severe class imbalances in real-world.

\noindent\textbf{Influence of BAM}. To assess the impact of the BAM design on model performance, we compare the TG-MILNet model with and without the BAM component. As illustrated in Table~\ref{table:dual}, the results demonstrate its contribution beyond a mere improvement in accuracy. Removing this component results in a 6.0\% drop in accuracy, indicating its essential role in handling borderline cases, particularly in distinguishing subtle spinal curvatures. The most significant decline is observed in specificity, which decreases by 8.0\% , highlighting BAM’s role in maintaining specificity and reducing false positives. While sensitivity remains high, it confirms that the model continues to effectively detect scoliosis cases. However, the F1-score improves by 5.9\%, further validating that BAM optimizes classification balance, ensuring the model does not favor one class over another.

\begin{table}[t]
    \centering
    \caption{Ablative results on the impact of the BAM and textual guidance. The lower part shows class-wise recall improvement compared to ScoNet-MT.}
    \label{table:dual}
    \normalsize
    \resizebox{0.85\textwidth}{!}{
        \begin{tabular}{lcccc}
            \hline
            \textbf{Method}    & \textbf{Accuracy(\%)} & \textbf{Sensitivity(\%)} & \textbf{Specificity(\%)} & \textbf{F1$-$scores(\%)} \\ 
            \hline
            TG$-$MILNet  & \textbf{89.9}  & \textbf{99.5}  & \textbf{86.4}   & \textbf{90.2} \\
            \textit{w/o} BAM & 83.9  & 99.0  & 78.4  & 84.3 \\
            \textit{w/o} Textual guidance  & 88.9  & 99.0  & 85.1  & 89.4 \\
            \hline
            \multicolumn{5}{c}{\textbf{Class-wise Recall Comparison}} \\
            \hline
            \textbf{Class} & \textbf{ScoNet-MT (\%)} & \textbf{TG-MILNet (\%)} & \textbf{Recall Gain} & \\
            Negative & 99.0 & \textbf{99.5} & +0.6\% & \\
            Positive & 85.2 & \textbf{91.2} & +6.0\% & \\
            Neutral  & 43.7 & \textbf{70.6} & +27.0\% & \\
            \hline
        \end{tabular}
    }
\end{table}

\noindent\textbf{Influence of textual guidance}. 
To evaluate the impact of textual guidance, we compare models with and without it. As shown in Table~\ref{table:dual}, the most significant decline is in specificity, which decreases from 86.4\% to 85.1\%. This indicates that textual guidance plays a crucial role in reducing false positives, helping the model distinguish between normal and abnormal gait patterns. While the overall improvement may appear modest, we emphasize that the guidance improves the F1 on borderline cases by 5.3\% (1:1:8) and up to 10\% (1:1:16), demonstrating its value in challenging scenarios. The decline in other metrics suggests that textual guidance provides domain knowledge, enabling the model to focus on gait features related to scoliosis, thereby enhancing performance and improving its ability to identify borderline cases.

\section{Conclusion}
\label{sec:Conclusion}

In this study, we propose TG-MILNet, a Text-Guided Multi-Instance Learning Network, to enhance scoliosis detection by effectively handling class imbalance and borderline cases. By integrating DTW-based clustering, BAM, and IBTA, our model extracts discriminative gait patterns while reducing false positives. The incorporation of textual guidance further refines feature learning, improving interpretability and classification performance. Experimental results show that TG-MILNet outperforms existing SOTA methods, ensuring reliable scoliosis screening. Moreover, unlike pervious methods, TG-MILNet maintains robust across imbalance ratios, demonstrating its adaptability to real-world clinical scenarios. These results suggest that TG-MILNet is a promising tool for non-invasive scoliosis detection and holds great potential for other medical applications involving imbalanced data tasks, such as rare disease classification and anomaly detection in clinical imaging.

\subsection{Acknowledgements} This work was partially supported by US National Science Foundation IIS-2412195, CCF-2400785, the Cancer Prevention and Research Institute of Texas (CPRIT) award (RP230363) and the National Institutes of Health (NIH) R01 award  (1R01AI190103-01).
\subsection{Disclosure of Interests} The authors have no competing interests to declare that
are relevant to the content of this article.

%
%
%
%

\end{document}